\providecommand{\paperdocumentclass}{article}
\providecommand{\paperclassoptions}{11pt}
\documentclass[\paperclassoptions]{\paperdocumentclass}

\providecommand{\papernatbiboptions}{square,numbers,sort&compress}

\newif\ifpaperpreprintlayout
\ifdefined\papervenuebuild
  \paperpreprintlayoutfalse
\else
  \paperpreprintlayouttrue
\fi

\usepackage[T1]{fontenc}
\usepackage[utf8]{inputenc}
\ifpaperpreprintlayout
  
\usepackage[letterpaper,left=0.96in,right=0.96in,top=1.10in,bottom=1.05in,headheight=22pt,headsep=20pt,footskip=30pt]{geometry}
\usepackage{XCharter}
\usepackage[xcharter]{newtxmath}
\usepackage[scaled=0.95]{sourcesanspro}
\usepackage{microtype}
\usepackage{setspace}
\usepackage{caption}
\usepackage{fancyhdr}
\usepackage{needspace}
\usepackage{ragged2e}
\usepackage{titlesec}
\usepackage{xurl}

\fi
\usepackage{graphicx}
\usepackage{booktabs}
\usepackage{tabularx}
\usepackage{threeparttable}
\usepackage{array}
\usepackage{xcolor}
\usepackage{enumitem}
\usepackage{csquotes}
\usepackage{etoolbox}
\usepackage{subfiles}
\expandafter\usepackage\expandafter[\papernatbiboptions]{natbib}
\usepackage[acronym,nonumberlist,nopostdot]{glossaries-extra}
\usepackage[hidelinks]{hyperref}
\usepackage{orcidlink}
\usepackage[capitalise,noabbrev]{cleveref}

\graphicspath{{paper/figures/generated/}}
\setabbreviationstyle[acronym]{long-short}
\glsdisablehyper
\definecolor{draftred}{HTML}{A61B1B}

\usepackage[section]{placeins}

\newcommand{\paperbrand}{}
\newcommand{\paperversion}{}
\newcommand{\papertitle}{PocketTSLM: Deploying an On-Device Soft-Prompt Time-Series Language Model for Privacy-Preserving Smartphone Reasoning}
\newcommand{\papersubtitle}{}
\newcommand{\paperpdftitle}{PocketTSLM: Deploying an On-Device Soft-Prompt Time-Series Language Model for Privacy-Preserving Smartphone Reasoning}
\newcommand{\lastauthororcid}{0000-0002-8607-9148}
\newcommand{\rosenblattlorcid}{0009-0007-6084-1441}
\newcommand{\millingorcid}{0000-0002-8842-2958}
\newcommand{\kaarorcid}{0009-0008-9287-3138}
\newcommand{\jakoborcid}{0000-0003-4793-1366}
\newcommand{\schullerorcid}{0000-0002-6478-8699}
\newcommand{\fleischorcid}{0000-0002-4842-1117}
\newcommand{\langerorcid}{0009-0000-6669-3928}
\newcommand{\paperauthors}{Max Rosenblattl\authororcid{\rosenblattlorcid}\textsuperscript{1}, Nikhil Krishnaswamy\textsuperscript{1}, Leon Nissen\,\orcidlink{0009-0009-3884-0850}\textsuperscript{1}, Manuel Milling\authororcid{\millingorcid}\textsuperscript{2}, Nicolas Zumarraga\textsuperscript{4}, Ning Wang\textsuperscript{4}, Thomas Kaar\authororcid{\kaarorcid}\textsuperscript{1,3,4}, Robert Jakob\authororcid{\jakoborcid}\textsuperscript{3,4}, Bj\"orn W.\ Schuller\authororcid{\schullerorcid}\textsuperscript{2,5}, Elgar Fleisch\authororcid{\fleischorcid}\textsuperscript{4,6}, Patrick Langer\authororcid{\langerorcid}\textsuperscript{1,3,4}, and Paul Schmiedmayer\authororcid{\lastauthororcid}\textsuperscript{1}}
\newcommand{\paperpdfauthors}{Max Rosenblattl, Nikhil Krishnaswamy, Leon Nissen, Manuel Milling, Nicolas Zumarraga, Ning Wang, Thomas Kaar, Robert Jakob, Björn W.\ Schuller, Elgar Fleisch, Patrick Langer, and Paul Schmiedmayer}
\newcommand{\paperaffiliations}{\textsuperscript{1}Stanford University, USA; \textsuperscript{2}CHI, Technical University of Munich University Hospital, Germany; \textsuperscript{3}Aionic Labs, Switzerland; \textsuperscript{4}ETH Zurich, Switzerland; \textsuperscript{5}GLAM, Imperial College London, UK; \textsuperscript{6}Centre for Digital Health Interventions (CDHI), University of St.\ Gallen, Switzerland}
\newcommand{\paperdate}{October 14, 2026}
\newcommand{\shorttitle}{PocketTSLM}

\newcommand{\correspondingemail}{mrosenbl@stanford.edu}
\newcommand{\paperrepositoryurl}{https://github.com/SchmiedmayerLab/PocketTSLM}
\newcommand{\paperrepositorylabel}{github.com/SchmiedmayerLab/PocketTSLM}
\newcommand{\papercodelicense}{MIT}
\newcommand{\paperdoiurl}{}
\newcommand{\paperdoilabel}{}
\newcommand{\paperkeywords}{}
\newcommand{\papersubject}{Preprint manuscript}
\newcommand{\papernote}{A version of this paper was accepted at the Workshop on Deploying AI in Healthcare (DAIH) at the Conference on Language Modeling (COLM 2026).}

\newif\ifpaperdraft
\paperdraftfalse

\newcommand{\doi}[1]{\href{https://doi.org/#1}{doi:#1}}
\newcommand{\authororcid}[1]{\ifdefempty{#1}{}{\nobreak\hspace{0.22em}\orcidlink{#1}}}

\newcommand{\paperstartmaintext}{}
\newcommand{\paperkeywordsblock}{\ifdefempty{\paperkeywords}{}{\par\medskip
    \noindent\textbf{Keywords:} \paperkeywords
  }}
\newcommand{\paperbibliographystyle}{unsrtnat}
\ifSubfilesClassLoaded{\newif\ifpaperstandalonecitations
  \expandafter\pretocmd\csname NAT@sort@cites\endcsname{\global\paperstandalonecitationstrue}{}{}\pretocmd{\nocite}{\global\paperstandalonecitationstrue}{}{}\AtEndDocument{\ifpaperstandalonecitations
      \bibliographystyle{\paperbibliographystyle}\bibliography{references}\fi
  }}{}

\ifpaperdraft
  \input{paper/setup/draft-commands}
\fi

\newacronym{BF16}{bf16}{bfloat16}
\newacronym{BPE}{BPE}{byte-pair encoding}
\newacronym{CI}{CI}{continuous integration}
\newacronym{CNN}{CNN}{convolutional neural network}
\newacronym{COT}{CoT}{chain-of-thought}
\newacronym{CPU}{CPU}{central processing unit}
\newacronym{ECG}{ECG}{electrocardiogram}
\newacronym{EEG}{EEG}{electroencephalography}
\newacronym{GDPR}{GDPR}{General Data Protection Regulation}
\newacronym{GPU}{GPU}{graphics processing unit}
\newacronym{HIPAA}{HIPAA}{Health Insurance Portability and Accountability Act}
\newacronym{IRB}{IRB}{institutional review board}
\newacronym{LLM}{LLM}{large language model}
\newacronym{LORA}{LoRA}{low-rank adaptation}
\newacronym{MLP}{MLP}{multilayer perceptron}
\newacronym{OPENTSLMSP}{OpenTSLM-SP}{OpenTSLM soft-prompt}
\newacronym{OS}{OS}{operating system}
\newacronym{QKV}{QKV}{query-key-value}
\newacronym{TSLM}{TSLM}{time-series language model}
\newacronym{TTFT}{TTFT}{time to first token}
\newglossaryentry{ECGQA}{name={ECG-QA},description={electrocardiogram question answering benchmark}}
\newglossaryentry{LLASA}{name={LLaSA},description={Large Language and Sensor Assistant}}
\newglossaryentry{MLX}{name={MLX},description={Apple machine learning framework}}
\newglossaryentry{MLXSWIFT}{name={MLX-Swift},description={Swift interface for MLX}}
\newglossaryentry{ODCSIS}{name={ODCSIS},description={On-Device Contextual Sensing and Inference System}}
\newglossaryentry{OPENTSLM}{name={OpenTSLM},description={OpenTSLM model}}
\newglossaryentry{SENSORLLM}{name={SensorLLM},description={SensorLLM model}}

\ifpaperpreprintlayout

\definecolor{paperink}{HTML}{2E2D29}
\definecolor{paperaccent}{HTML}{8C1515}
\definecolor{paperaccentdark}{HTML}{820000}
\definecolor{papermuted}{HTML}{53565A}
\definecolor{paperrule}{HTML}{C0C0BF}
\definecolor{paperwash}{HTML}{EAEAEA}

\setlist{leftmargin=*,topsep=0.5em,itemsep=0.22em,parsep=0pt,partopsep=0pt}

\titleformat{\section}
  {\bfseries\fontsize{15}{18}\selectfont\color{paperink}}
  {\thesection}
  {0.72em}
  {}
\titleformat{\subsection}
  {\bfseries\fontsize{12.2}{15}\selectfont\color{paperink}}
  {\thesubsection}
  {0.66em}
  {}
\titleformat{\subsubsection}
  {\bfseries\normalsize\color{paperink}}
  {\thesubsubsection}
  {0.62em}
  {}
\titleformat{\paragraph}[runin]
  {\bfseries\color{paperink}}
  {}
  {0pt}
  {}
  [.\quad]
\titlespacing*{\section}{0pt}{1.75\baselineskip plus 0.30\baselineskip minus 0.20\baselineskip}{0.52\baselineskip}
\titlespacing*{\subsection}{0pt}{1.20\baselineskip plus 0.20\baselineskip minus 0.15\baselineskip}{0.32\baselineskip}
\titlespacing*{\subsubsection}{0pt}{0.95\baselineskip plus 0.15\baselineskip minus 0.10\baselineskip}{0.24\baselineskip}
\titlespacing*{\paragraph}{0pt}{0.8\baselineskip}{0.5em}
\pretocmd{\section}{\Needspace{5\baselineskip}}{}{}
\makeatletter
\pretocmd{\subsection}{\if@nobreak\else\Needspace{4\baselineskip}\fi}{}{}
\makeatother

\newcommand{\preprintseparator}{\hspace{0.62em}\textcolor{paperrule}{\textperiodcentered}\hspace{0.62em}}
\newcommand{\preprintdateline}{\sffamily\fontsize{8.7}{10.5}\selectfont\color{papermuted}\ifdefempty{\paperbrand}{}{\paperbrand\preprintseparator}\ifdefempty{\paperversion}{}{\paperversion}\ifdefempty{\paperdate}{}{\ifdefempty{\paperversion}{}{\preprintseparator}\paperdate}}
\newif\ifpreprintmetadata

\renewcommand{\paperstartmaintext}{\clearpage}

\renewcommand{\maketitle}{\thispagestyle{preprintfirst}\vspace*{0.15in}\begingroup
  {\raggedright\bfseries\fontsize{25}{28.5}\selectfont\color{paperink}\papertitle\par}
  \ifdefempty{\papersubtitle}{}{\vspace{0.42em}
    {\raggedright\fontsize{13}{16}\selectfont\color{papermuted}\papersubtitle\par}}
  \vspace{0.95em}
  {\raggedright\bfseries\fontsize{11.1}{13.6}\selectfont\color{paperink}\paperauthors\par}
  \ifdefempty{\paperaffiliations}{}{\vspace{0.38em}
    {\raggedright\fontsize{9.2}{11.6}\selectfont\color{papermuted}\paperaffiliations\par}}
  \preprintmetadatafalse
  \vspace{0.64em}
  {\raggedright\sffamily\fontsize{8.45}{10.6}\selectfont\color{papermuted}\ifdefempty{\correspondingemail}{}{\textbf{Correspondence}\enspace\href{mailto:\correspondingemail}{\nolinkurl{\correspondingemail}}\preprintmetadatatrue
    }\ifdefempty{\paperrepositoryurl}{}{\ifpreprintmetadata\preprintseparator\fi
      \textbf{Code}\enspace\href{\paperrepositoryurl}{\nolinkurl{\paperrepositorylabel}}\preprintmetadatatrue
    }\ifdefempty{\papercodelicense}{}{\ifpreprintmetadata\preprintseparator\fi
      \textbf{Code license}\enspace\papercodelicense
      \preprintmetadatatrue
    }\ifdefempty{\paperdoiurl}{}{\ifpreprintmetadata\preprintseparator\fi
      \textbf{DOI}\enspace
      \href{\paperdoiurl}{\ifdefempty{\paperdoilabel}{DOI}{\nolinkurl{\paperdoilabel}}}\preprintmetadatatrue
    }\par
  }
  \endgroup
}

\renewenvironment{abstract}
  {\par\vspace{1.10em}
    \noindent\textcolor{paperrule}{\rule{\textwidth}{0.5pt}}
    \par\vspace{0.67em}
    \noindent{\bfseries\fontsize{11.2}{13.5}\selectfont\color{paperink}Abstract\par}
    \vspace{0.24em}
    \begingroup
    \small
    \setstretch{1.065}
    \setlength{\parindent}{0pt}
    \setlength{\parskip}{0.34em}}
  {\par
    \endgroup
    \vspace{0.58em}
    \noindent\textcolor{paperrule}{\rule{\textwidth}{0.5pt}}
    \par\vspace{1.08em}
  }

\renewcommand{\paperkeywordsblock}{\ifdefempty{\paperkeywords}{}{\par\vspace{0.58em}
    {\sffamily\footnotesize\color{papermuted}\textbf{\color{paperink}Keywords}\enspace\paperkeywords\par}}}

\fancypagestyle{preprint}{\fancyhf{}
  \fancyhead[L]{\sffamily\footnotesize\color{papermuted}\shorttitle}
  \fancyfoot[C]{\sffamily\footnotesize\color{papermuted}\thepage}
  \renewcommand{\headrulewidth}{0.4pt}
  
}
\fancypagestyle{preprintfirst}{\fancyhf{}
  \fancyhead[R]{\preprintdateline}
  \fancyfoot[C]{\sffamily\footnotesize\color{papermuted}\thepage}
  \renewcommand{\headrulewidth}{0.4pt}
  
}
\fancypagestyle{plain}{\fancyhf{}
  \fancyhead[L]{\sffamily\footnotesize\color{papermuted}\shorttitle}
  \fancyfoot[C]{\sffamily\footnotesize\color{papermuted}\thepage}
  \renewcommand{\headrulewidth}{0.4pt}
  
}
\renewcommand{\headrule}{\hbox to\headwidth{\color{paperrule}\leaders\hrule height \headrulewidth\hfill}}

\hypersetup{
  colorlinks=true,
  linkcolor=paperaccentdark,
  citecolor=paperink,
  urlcolor=paperaccentdark,
  pdfborder={0 0 0},
  pdftitle={\paperpdftitle},
  pdfauthor={\paperpdfauthors},
  pdfsubject={\papersubject},
  pdfkeywords={\paperkeywords}
}

\AtBeginDocument{\color{paperink}
  \pagestyle{preprint}
}

\AtBeginEnvironment{thebibliography}{\clearpage\thispagestyle{preprint}\interlinepenalty=10000}

\fi

\title{\papertitle}
\author{\paperauthors}
\date{\paperdate}

\begin{document}

\maketitle
\ifdefempty{\papernote}{}{{\renewcommand{\thefootnote}{}\begin{NoHyper}\footnotetext{\papernote}\end{NoHyper}}}

\begin{abstract}
Consumer phones and wearables increasingly collect physiological time-series data, yet most mobile systems treat these streams as summaries or text rather than as temporal evidence.
This representation is poorly matched to personal health systems that need to retrieve, route, or summarize sensor evidence while preserving temporal structure.
Cloud-hosted time-series language models avoid smartphone limits, but they repeatedly move sensitive physiology off the device and add latency and per-query compute cost.
Mobile model alternatives that aggregate time-series data into statistical features or serialize raw samples as text either compress temporal structure or exhaust the context and memory budgets of small on-device models.
We present PocketTSLM, a mobile-native deployment of OpenTSLM SoftPrompt, the soft-prompt variant of the OpenTSLM time-series language model.
PocketTSLM runs fully on a consumer smartphone by injecting encoded time-series windows as embeddings instead of tokenizing time series as text.
A numerical parity protocol verifies identical greedy-decoded token sequences between the Swift port and a Python reference using Apple's MLX machine-learning framework.
Answers also match the PyTorch reference, although rationale wording can differ.
An on-device context-economics benchmark compares the soft-prompt path with a text-serialized baseline on the same backbone.
In our benchmark format, soft-prompting uses approximately $12\times$ fewer time-series tokens and reduces prompt-prefill cost.
PocketTSLM processes 30-second Sleep-EDF electroencephalography clips for sleep staging and a 1{,}000-sample (10\,s) 12-lead electrocardiogram clip from ECG-QA.
In comparison, text-serialized baselines exceed the phone's context or memory budget at longer time-series windows.
All inference runs entirely on the device, with no health data transmitted during inference.
\paperkeywordsblock
\end{abstract}

\glsresetall

\paperstartmaintext

\section{Introduction}

Consumer devices now collect physiological time-series data that were previously limited to clinical or research environments~\citep{steinhubl2015mobile, li2017physiomes}.
Smartwatch irregular-pulse notifications have been evaluated for atrial fibrillation screening in large cohorts~\citep{appleheart2019}, phone-connected systems make \gls{EEG} acquisition more accessible~\citep{bauer-eeg}, and passive wearable streams such as heart rate, sleep, and activity have been used for infectious-disease surveillance, COVID-19 detection, and prospective health-risk assessment~\citep{radin2020influenza, quer2021covid, strain2020futurehealth}.
In each case, measurements are captured close to the patient, but the interpretation delivered to a patient or downstream system is often a label, score, or study endpoint.
Digital biomarker pipelines make this pattern explicit: continuous measurements are translated into digital endpoints or biomarkers for evaluation and decision support~\citep{coravos2019digitalbiomarkers}.
Those outputs support screening, monitoring, and study endpoints, but they do not explain the source time-series window itself.
Recent personal-health \gls{LLM} systems interpret wearable-derived sleep, activity, and fitness data through natural-language coaching or insight generation~\citep{khasentino2025personalhealth, merrill2024phia}.
Therefore, patient-facing assistants, clinician review tools, and supervised health agents need descriptions grounded in source time-series windows that can be retrieved, routed, summarized, or used to defer a response when the evidence is insufficient.

\Glspl{TSLM} are designed for this time-series-to-language interface.
They couple time-series encoders to language models and generate natural-language descriptions or rationales from raw physiological time-series data such as \gls{ECG}, \gls{EEG}, and accelerometry~\citep{opentslm, xu2026ecg}.
These outputs give software a text interface to sensor evidence while preserving the temporal structure in the model input~\citep{opentslm}.
Time Series Language Models may enable systems that reason over sensor windows rather than only aggregate features.

Personal-health \gls{LLM} systems are important reference points, but they do not solve the deployment problem addressed here: they reason over derived wearable data with large language models rather than running a time-series-native model on the phone~\citep{khasentino2025personalhealth, merrill2024phia}.
When interpretation is served remotely, personal physiological streams or derived sensor evidence must leave the device, making privacy and secondary-use risks part of the inference pathway~\citep{price2019privacy}.
Server-side inference also adds a network dependency and per-query compute cost.
General on-device \gls{LLM} work identifies limited device memory as a central inference bottleneck~\citep{alizadeh2024llmflash}.
Medical \gls{LLM} benchmarks show the same tradeoff: local inference improves the privacy and cost posture, while memory and efficiency become the limiting constraints~\citep{medicine-edge}.
For physiological time-series data, moving inference onto the phone shifts the system boundary: raw data can remain local, offline analysis becomes possible, and cloud storage is no longer required for each query.

The main obstacle is representation rather than model size.
Mobile \gls{LLM} runtimes are built around token-identifier input and do not expose a path for injecting precomputed embeddings, so a time-series encoder's output cannot be fed to the model directly.
If the runtime cannot accept encoder embeddings, the time-series data must be serialized as text.
Text serialization reuses an unchanged language-model interface and can cast time series as next-token prediction~\citep{gruver2023llmtime}, but it expands each sample into multiple vocabulary tokens, formats or rounds numeric values, and spends the same limited context window needed for instructions and generated text.
\Gls{OPENTSLMSP} avoids that prompt-level bottleneck by using representation-level fusion: a time-series encoder projects time-series windows into soft-prompt embeddings that are injected directly into the \gls{LLM}'s embedding sequence~\citep{opentslm}.
That design is better matched to physiological time-series data, but it requires embedding-level access to the forward pass and cannot be expressed through token-only generation interfaces.

We present PocketTSLM, an on-device port of \gls{OPENTSLMSP} to \gls{MLXSWIFT} that evaluates whether the soft-prompt pathway can run on a consumer smartphone without changing the reference model's greedy-decoding behavior.
The port preserves the encoder-projector-\gls{LLM} architecture while modifying the on-device generation path to accept projected time-series embeddings.
PocketTSLM is scoped as a deployment and evaluation of on-device soft-prompt physiological time-series reasoning, not clinical validation, autonomous medical action, or behavior-change efficacy.
We contribute: \textbf{(1)} a method for migrating soft-prompt \glspl{TSLM} to \gls{MLXSWIFT}; \textbf{(2)} a numerical parity protocol that compares the on-device implementation against the reference under greedy decoding; \textbf{(3)} an on-device context-economics benchmark comparing soft-prompt encoding with a text-serialized baseline on the same backbone; and \textbf{(4)} a HealthKit-enabled iPhone demo that runs \gls{ECG} and \gls{EEG} reasoning locally without sending health data off-device.

\section{Related Work}

\paragraph{Time-series interfaces for language models}
Integrating time-series data into \glspl{LLM} is an active research area \citep{tsllm-survey}, with approaches spanning text serialization, model reprogramming, soft prompting, and cross-attention fusion \citep{alayrac2022}.
Text serialization preserves the ordinary token interface and has been used for forecasting and wearable-health prediction \citep{gruver2023llmtime, health-llm}, but language models often struggle to reason over numbers presented this way \citep{merrill2024}.
Since every serialized sample consumes vocabulary tokens, long time-series windows can spend the same context window needed for instructions and generated text.
Reprogramming methods keep the language-model backbone frozen while aligning time-series patches or text prototypes to the model input space \citep{jin2024timellm, sun2023test}.
These methods show that representation choice matters, but they primarily target prediction, classification, or forecasting rather than natural-language rationales grounded in raw time-series data \citep{gruver2023llmtime, health-llm, jin2024timellm, sun2023test}.
Representation-level fusion instead learns a time-series encoder whose embeddings are injected into the model as soft prompts or cross-attention inputs, preserving free-form generation \citep{chow2024, chatts}.
\gls{OPENTSLM} \citep{opentslm} uses this approach for \gls{COT} reasoning \citep{wei2022cot}, coupling a time-series encoder to a frozen \gls{LLM} through soft prompts and generating rationales for clinical time-series tasks.
We port its soft-prompt variant fully onto a smartphone and quantify the token, memory, and latency cost of soft prompts against text serialization on the same backbone (\S\ref{sec:results}).

\paragraph{Wearable and physiological time-series language systems}
In personal-health and wearable sensing, recent \gls{LLM} systems often reason over derived variables, contextual summaries, or tool-generated features rather than raw physiological time-series windows \citep{health-llm, merrill2024phia}.
\gls{SENSORLLM} and \gls{LLASA} align inertial measurements with language outputs for human activity recognition and sensor question answering, but they focus on motion sensing rather than physiological \gls{ECG} or \gls{EEG} interpretation \citep{li2025sensorllm, imran2024llasa}.
\gls{ECGQA} provides a benchmark for \gls{ECG} question answering, and ECG-Chat trains an \gls{ECG}-language model for report generation and dialogue \citep{oh2023ecgqa, zhao2024ecgchat}.
These systems expand time-series language modeling for health and wearables, but they do not establish whether a soft-prompt physiological \gls{TSLM} can run fully on a consumer smartphone.

\paragraph{Deployment settings for health time-series inference}
Most time-series-to-language and wearable-health language models are developed and evaluated in server-side or offline research settings \citep{opentslm, health-llm, merrill2024phia, zhao2024ecgchat}, so privacy, latency, and per-query cost remain deployment constraints rather than solved properties of the model.
On-device \gls{LLM} inference is an emerging direction, but it is tightly constrained by memory and efficiency, including for clinical models \citep{alizadeh2024llmflash, medicine-edge}.
For privacy, federated learning decentralizes training rather than inference \citep{mcmahan2017}, and fog execution keeps inference within a trusted local network \citep{spezillm-fog}.
On-device, \gls{ODCSIS} \citep{an2025odcsis} runs a small language model on a phone for human activity recognition, but injects sensors by serializing statistical features into text rather than encoding the raw time-series data.
PocketTSLM therefore targets a remaining deployment constraint: soft-prompt \glspl{TSLM} require embedding-level input and enough context headroom for raw time-series windows, whereas text serialization preserves an unchanged token-level generation path at the cost measured in \S\ref{sec:results}.
Prior work therefore covers time-series-to-language modeling, wearable and physiological language systems, privacy-preserving inference, and on-device language models, but not soft-prompt physiological \gls{TSLM} inference fully on a consumer smartphone.

\section{Methods}
\label{sec:method}

\paragraph{Reference model and migration strategy}
PocketTSLM ports \gls{OPENTSLMSP}, the soft-prompt variant of \gls{OPENTSLM}, from PyTorch into an iOS/\gls{MLXSWIFT} runtime.
The PyTorch reference does not run in the target environment without reimplementation, as the model depends on framework-specific modules and a generation path that accepts projected input embeddings.
We migrate the model in two stages: first from PyTorch to Python \gls{MLX}, then from Python \gls{MLX} to Swift \gls{MLX}.
This staging separates framework-level parity from Swift runtime parity and provides a Python \gls{MLX} reference for the final port.

\begin{figure}[t]
  \centering
  \includegraphics[width=\textwidth,alt={On-device architecture: physiological signal patches pass through a time-series encoder and projector, then combine with text embeddings for a language model with a LoRA adapter to generate a rationale.}]{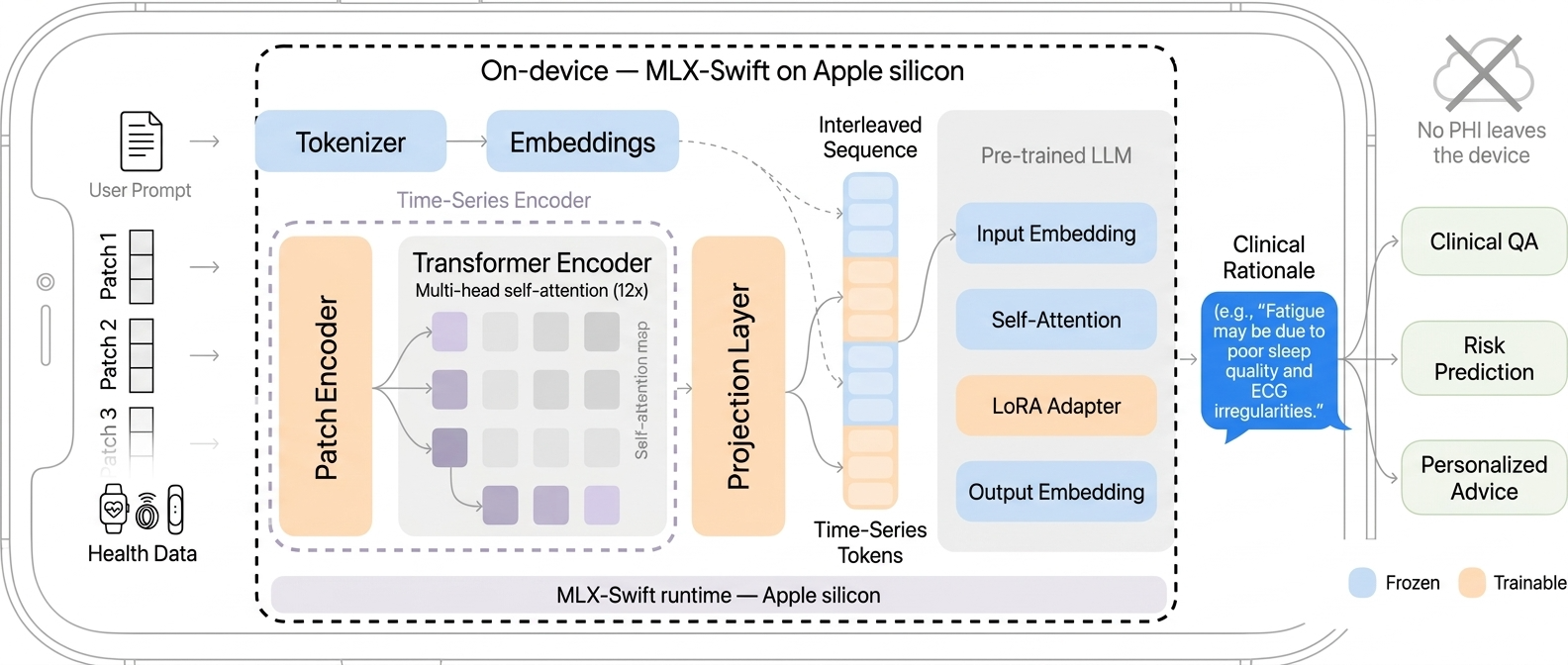}
  \caption{\textbf{The \gls{OPENTSLMSP} pipeline running fully on-device.}
  Health time-series data are split into patches and encoded by a \gls{CNN} patch encoder and transformer (the time-series encoder), projected into the \gls{LLM} embedding space, and interleaved with the embedded text prompt.
  The combined sequence is decoded by a frozen pre-trained \gls{LLM} with a \gls{LORA} adapter into a natural-language model rationale.
  The entire pipeline executes on the device through the \gls{MLXSWIFT} runtime on Apple silicon, with no health data leaving the phone.
  Figure adapted from \citet{opentslm}.}
  \label{fig:pipeline}
\end{figure}

\paragraph{Soft-prompt architecture}
The architecture receives channel- or lead-wise physiological time-series windows, including the single-channel \gls{EEG} (Sleep-EDF sleep staging) and 12-lead \gls{ECG} (\gls{ECGQA}) inputs evaluated in \S\ref{sec:results}.
A raw physiological time series is split by temporal position into fixed-length patches and fed to a time-series encoder: a \gls{CNN} patch encoder followed by a Transformer that produces one embedding per patch (Fig.~\ref{fig:pipeline}).
Patch length, stride, z-normalization, and lead ordering follow the \gls{OPENTSLM} reference without modification.

A learned \gls{MLP} projector maps the encoder outputs into the input space of a frozen pre-trained \gls{LLM}, Llama-3.2-1B in our deployment.
The projected embeddings are interleaved with the embedded text prompt as \emph{soft-prompt tokens}, meaning continuous vectors are injected directly into the model's embedding sequence rather than vocabulary tokens. \citep{opentslm}
Different \gls{LORA} adapters~\citep{lora}, each roughly 1\% of the model's parameters, adapt the frozen backbone to the \gls{ECG} and \gls{EEG} tasks.
The selected adapter determines the modality-specific mapping used at inference.
This representation-level pathway reduces prompt length, as each time-series patch becomes one embedding, but it requires the decoder forward pass to accept precomputed embeddings rather than only token identifiers.

\paragraph{Migration implementation}
The first migration stage reimplements three \gls{OPENTSLMSP} components in Python \gls{MLX}: the \texttt{TransformerCNNEncoder}, the \texttt{MLPProjector}, and the interleave logic that inserts projected time-series embeddings between text-token embeddings before the \gls{LLM} forward pass.
The trained encoder, projector, and \gls{LORA} adapter weights are loaded directly from the original PyTorch checkpoints and converted to the \gls{MLX} format.
Conversion applies layout transforms where the two frameworks use different weight conventions, including Conv1d kernel transposition and fused \gls{QKV} splitting.
The \gls{LLM} backbone uses Llama-3.2-1B with \gls{BF16} base weights, matching the adapted checkpoint format.

The second migration stage ports the Python \gls{MLX} reference module by module to Swift \gls{MLX}, targeting Apple's \gls{MLXSWIFT} framework.
Since both stages share \gls{MLX} operator semantics, discrepancies introduced in the Swift port can be separated from discrepancies introduced during the PyTorch-to-\gls{MLX} conversion.
\Gls{MLXSWIFT}'s high-level generation path could not be used directly, as it accepts token identifiers rather than precomputed input embeddings.
\Gls{MLXSWIFT} exposes the lower-level tensor operations needed to rebuild the pipeline, so we adapted its Llama implementation and forward pass to accept input embeddings.
Since the Swift modules use the same parameter keys as the Python \gls{MLX} reference, the encoder and projector weights load from \texttt{safetensors} without further remapping.

Three implementation details affect reproducibility.
Several NumPy-style convenience functions available in Python \gls{MLX} have no direct Swift counterpart and were reconstructed from lower-level \texttt{MLXArray} operations.
\Gls{MLX}'s lazy-evaluation model requires explicit evaluation barriers during the encoder-projector hand-off so intermediate tensors are materialized before generation.
Certain operations dispatch to Metal \gls{GPU} kernels unavailable in headless \gls{CI} environments, so parity tests are pinned to the \gls{CPU} backend for deterministic comparison.

\paragraph{Evaluation data and benchmark protocol}
The reported benchmarks use fixed, public, de-identified research samples from Sleep-EDF \citep{kemp2000sleepedf, goldberger2000physionet} and \gls{ECGQA} \citep{oh2023ecgqa}; they do not use prospective HealthKit records from patients.
We evaluate two \gls{OPENTSLM} tasks: single-channel Sleep-EDF sleep staging and 12-lead \gls{ECGQA}.
Both tasks run with a Llama-3.2-1B backbone, \gls{BF16} base weights, and the original \gls{LORA} checkpoints without retraining.

The benchmark has two components.
First, a numerical parity protocol compares the Swift \gls{MLX} port with the reference implementation under greedy decoding.
For each evaluated prompt, the protocol compares the full greedy-decoded token sequence from the Swift port against the Python \gls{MLX} and PyTorch references.

Second, a context-economics benchmark compares two encodings of the same raw time series with the same backbone and generation loop.
Text serialization is the natural baseline for soft prompting on device: it preserves an unmodified token-level generation interface and is the established method for feeding raw time series to a language model as tokens \citep{gruver2023llmtime, health-llm}.

The soft-prompt arm injects representation-level embeddings and produces $\lceil L/4 \rceil$ temporal tokens per channel or lead, where $L$ is the length of the time-series.
The text-serialized arm uses \gls{OPENTSLM}'s prompt-level format~\citep{opentslm}, a fixed-precision encoding in the style of LLMTime~\citep{gruver2023llmtime}: each z-normalized sample is rounded to two decimals, its decimal point dropped, and the result rendered as \texttt{"012"}-style text fed as vocabulary tokens.
For the zero-padded sample format used here, the text arm costs 3 \gls{BPE} tokens per raw sample per channel or lead, measured with the backbone tokenizer.
Both arms decode greedily with a maximum of 80 new tokens, so differences in prompt tokens, \gls{TTFT}, and peak memory are attributable to the input encoding rather than the decoder policy.

We sweep series length $L$ and report, per encoding, time-series tokens, full-prompt feasibility, \gls{TTFT}, peak memory, and the context-window ceiling.
We report \gls{TTFT} rather than end-to-end latency, as truncating below the training window can change the generated-token count, while the prefill cost is directly driven by the input encoding.
The theoretical capacity estimates use the backbone's 8192-token context window, whereas the measured on-device sweep uses a lower 4,300-token prompt budget to preserve runtime memory headroom during generation.
The two arms share the same \gls{LORA}-adapted backbone.
\Gls{LORA} changes generated content but not prefill cost or memory, so the reported context economics are unaffected by the adapter choice.
All on-device measurements were collected on an iPhone Air running iOS~26.4.1 with \gls{MLXSWIFT}~0.21.3.
The reported \gls{TTFT} and peak memory values are means over five repeated sweeps of the same configuration, given as mean $\pm$ standard deviation.
\Gls{TTFT} is measured from the start of prompt prefill to the emission of the first generated token.
Peak memory is the high-water mark of \gls{MLX}'s unified-memory allocator, reset immediately before each arm, so it covers the resident model weights plus that arm's prefill activations and KV cache but excludes allocations made outside the \gls{MLX} allocator.

\begin{figure}[htbp]
  \centering
  \includegraphics[width=0.88\textwidth,alt={Three PocketTSLM app screens on an iPhone in airplane mode show the task menu, an EEG sleep-staging prompt, and the generated rationale and answer.}]{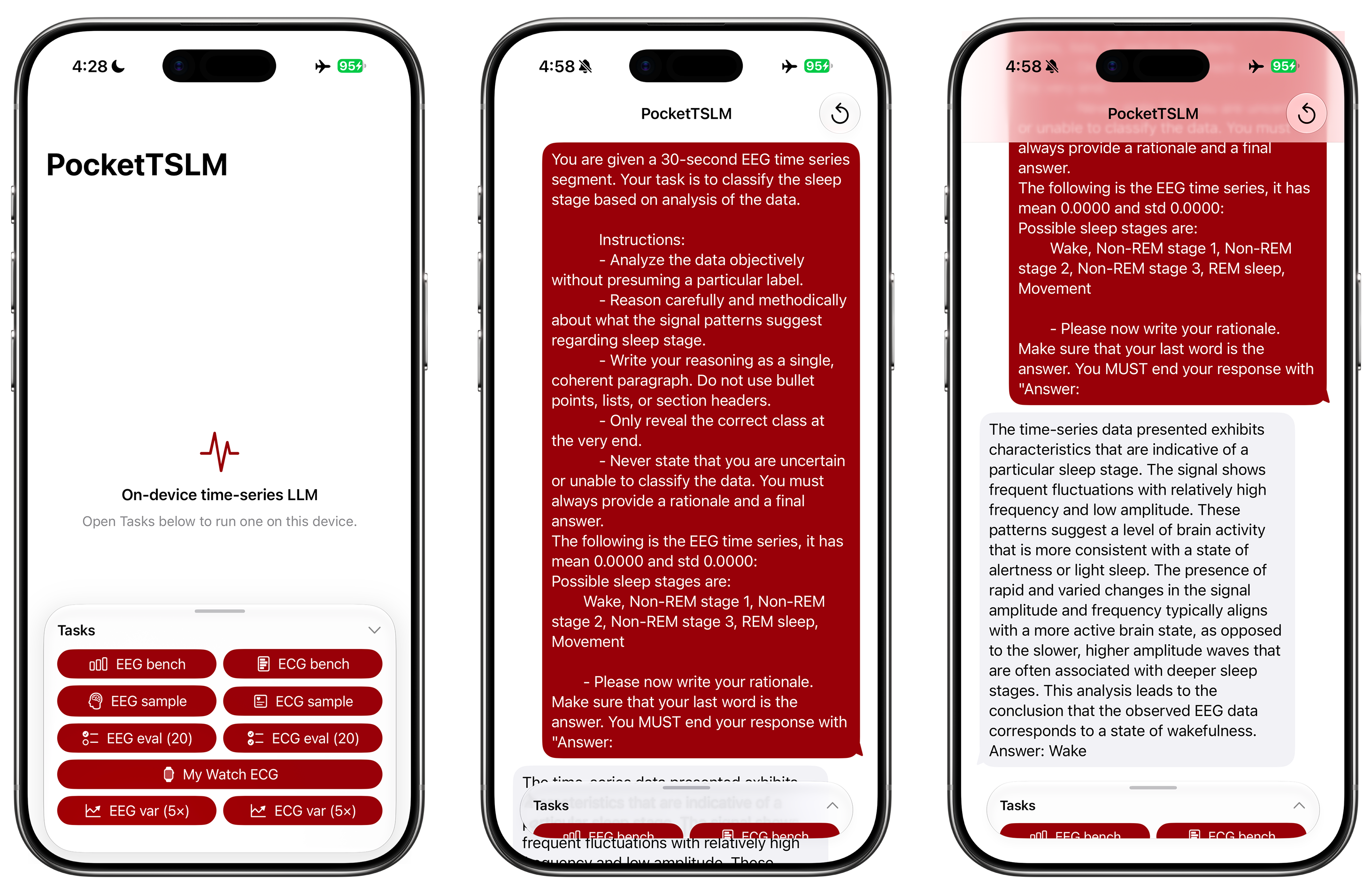}
  \caption{\textbf{On-device \gls{EEG} sleep staging in the demo app}, running on an iPhone Air in airplane mode.
  The app's task menu (left), the Sleep-EDF prompt with its \gls{EEG} input summary (center), and the model-generated rationale and answer (right).}
  \label{fig:app}
\end{figure}

\paragraph{App integration}
The on-device executable is embedded in a demo iOS app for local physiological time-series interpretation (Fig.~\ref{fig:app}).
The demo runs the pipeline on bundled evaluation samples, the same fixed samples used in the benchmark (\S\ref{sec:results}), included in the app bundle, and on request on the most recent \gls{ECG} recorded by a paired Apple Watch and read from HealthKit on the device.
All inference runs locally: the encoder, projector, and \gls{LLM} execute in-process through \gls{MLXSWIFT}, and no health data leaves the device.
Network access is unused during inference; the app makes no outgoing connections while running the pipeline.

\section{Results}
\label{sec:results}

These results are scoped to implementation parity and on-device context economics.
They do not test clinical performance or patient-facing deployment.

The benchmark protocol is defined in \S\ref{sec:method}; the results below report the observed token, latency, and memory consequences of the two encodings.
Prior clinician review of \gls{ECG} rationales from the \gls{OPENTSLM} model family supports the underlying task \citep{opentslm}, but the present study only tests the mobile port and benchmark behavior.

\paragraph{Implementation parity}
Four components (encoder, projector, fused encoder-projector pipeline, and rotary embedding) match the Python \gls{MLX} reference within a maximum element-wise tolerance of $10^{-4}$ and mean $10^{-5}$ ($10^{-3}$ for the rotary embedding), checked on the \gls{CPU} backend.
The tokenizer reproduces the reference token IDs exactly, and the text-serialization baseline is byte-identical to the \gls{OPENTSLM} format.
End-to-end, the Python \gls{MLX} and Swift implementations produce identical greedy-decoded token sequences.
Against the PyTorch reference the decoded answers agree, while the intervening rationale wording can differ under \gls{BF16} arithmetic.

\paragraph{On-device task performance}
The parity result above concerns implementation equivalence rather than task accuracy, so we also measure accuracy directly on device.
We pinned the first 20 samples of the Sleep-EDF \gls{COT} test split as a fixed export, so that index $i$ denotes the same recording in every implementation, and ran them under greedy decoding through the PyTorch reference, our Python \gls{MLX} reference, and the Swift port on the iPhone.
All three score $17/20$ and agree on every prediction, including the same three errors with the same incorrect labels, and the on-device prompt length matched the PyTorch reference at every sample.
Agreement on the errors is the stronger signal, since a port that only approximated the reference would diverge first where the model is least confident; it anchors the parity claim above to the original PyTorch implementation rather than only to our intermediate \gls{MLX} reference.
The measured $85\%$ is consistent with the $81.08\%$ \citet{opentslm} report on the full test split, but 20 samples give a $95\%$ interval of roughly $\pm16$ points, so this corroborates the deployment rather than estimating accuracy.
Soft prompting also does not trade quality for context economy: on this task the reference reports $69.88$ F1 against $9.05$ for the same backbone fine-tuned on a text serialization of the same signal \citep{opentslm}.
The compression ratio is fixed rather than tuned: patch length and stride follow the \gls{OPENTSLM} reference, as the encoder, projector, and \gls{LORA} adapters are loaded from its checkpoints, so varying it would require retraining all three, which is a change to the model rather than to its deployment.

\begin{figure}[!htbp]
  \centering
  \includegraphics[width=0.92\textwidth,alt={Two plots compare text serialization and soft prompting for EEG and ECG inputs across series lengths: time-series token counts on the left and time to first token on the right.}]{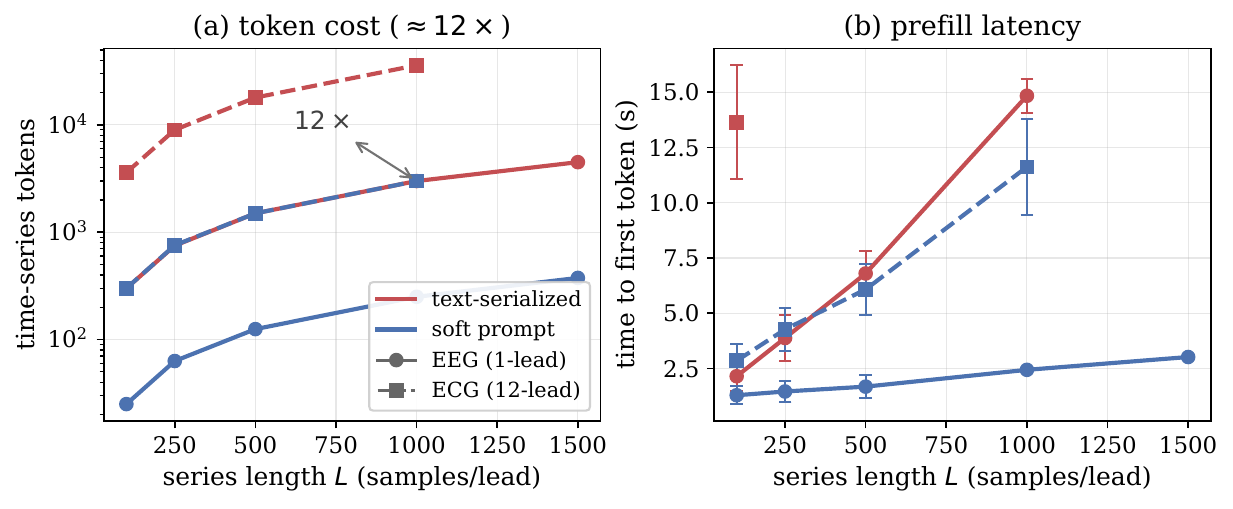}
  \caption{On-device context economics (iPhone, Llama-3.2-1B, greedy decoding), for single-channel \gls{EEG} (Sleep-EDF \citep{kemp2000sleepedf, goldberger2000physionet}) and 12-lead \gls{ECG} (\gls{ECGQA} \citep{oh2023ecgqa}).
  \textbf{(a)} Time-series tokens vs.\ series length: text serialization costs approximately $12\times$ as many tokens as soft prompts across evaluated lengths and both tasks (soft-prompt tokens are injected embeddings, not vocabulary tokens).
  \textbf{(b)} \Gls{TTFT}: soft-prompt \gls{TTFT} is lower across the evaluated sweep, while the text arm grows steeply and drops out once its full prompt exceeds the 4,300-token on-device budget ($L \ge 250$ for \gls{ECG}, $L \ge 1500$ for \gls{EEG}).
  Error bars show $\pm$ SD over five sweeps.
  Peak memory (Tables~\ref{tab:economics},~\ref{tab:economics_ecg}) tracks the same token count.}
  \label{fig:economics}
\end{figure}

\paragraph{Context economics}
Soft prompts produce one token per four-sample patch, rounding up partial patches, compared with 3 \gls{BPE} tokens per raw sample for text serialization in the benchmark format (Tables~\ref{tab:economics},~\ref{tab:economics_ecg}; Fig.~\ref{fig:economics}a).
This yields an approximately $12\times$ token advantage across evaluated lengths and both tasks, with rounding to whole patches giving $11.9\times$ at $L{=}250$.
Both encodings are charged per lead, so channel count preserves the per-lead token ratio.
A 12-lead \gls{ECG} window of $N$ samples costs $36N$ text tokens versus $12\lceil N/4 \rceil$ soft-prompt tokens.
Channel count grows the absolute token saving (12 leads $\times$ the per-lead gap).
Extrapolated to the backbone's 8192-token context window, text supports $\sim$2{,}659 single-channel \gls{EEG} samples ($\sim$53\,s of $50$\,Hz time-series data) but only $\sim$212 samples per \gls{ECG} lead, whereas soft prompts support $\sim$31{,}912 and $\sim$2{,}553 respectively under the same budget.
In the measured sweep, the lower 4,300-token on-device prompt budget is the practical cutoff used to preserve runtime memory headroom during generation.

\paragraph{Latency and memory}
The token gap dominates on-device cost, and the decisive case is 12-lead \gls{ECG}.
There, text serialization is infeasible beyond the shortest window.
At $L{=}100$ it already needs $4{,}133$ prompt tokens and $4.3$\,GiB, and every window extension ($L \ge 250$, $\ge 9{,}533$ tokens) is skipped to avoid an uncatchable \gls{OS} memory termination.
Soft prompts run the full evaluated $1{,}000$-sample clip with \gls{TTFT} $\le 11.6$\,s and peak memory $\le 4.2$\,GiB (Table~\ref{tab:economics_ecg}, Fig.~\ref{fig:economics}b).
The \gls{BF16} backbone weights alone account for about 2.3\,GiB of peak memory, and the remainder grows with prompt length.
At truncated lengths the soft-prompt arm's \emph{generated text} degenerates, since a window shorter than the $1{,}000$-sample training clip is out of distribution; the reported token, latency, and memory costs are unaffected, as they depend on the input encoding rather than on the decoded content.
\Gls{TTFT} and peak memory are means over five repeated sweeps.
For \gls{EEG}, the spread is under $6\%$ of the mean at $L\ge1000$ but reaches roughly $30\%$ at $L\le500$, where short prefills are dominated by fixed overhead rather than by compute.
For \gls{ECG}, it is $19$--$27\%$ at every length, driven by a faster first run.
For \gls{EEG}, the reported thermal state rises from \texttt{nominal} to \texttt{serious} within the first sweep and remains there.
For \gls{ECG}, it reads \texttt{nominal} at each run start, although the first run is $1.6\times$ faster than the last.
Thermal throttling is a possible explanation for the \gls{ECG} slowdown, but was not directly established.
The reported means include all five sweeps, including the faster first run, and therefore describe repeated-run performance rather than an isolated cold start or a verified steady state.
We do not report energy consumption.

Single-channel \gls{EEG} shows the same mechanism more gently: text survives to $L{=}1000$, but it prefills $\sim$7$\times$ the tokens of the soft-prompt arm, so its \gls{TTFT} is roughly $6\times$ larger ($14.8 \pm 0.8$\,s vs.\ $2.4 \pm 0.1$\,s at $L{=}1000$), and it still exceeds the budget at the full original $1{,}500$-sample window; soft-prompt \gls{TTFT} stays $\le 3.0$\,s and peak memory $\le 3.9$\,GiB at every length.

\begin{table}[htbp]
  \centering
  \footnotesize
  \caption{On-device context economics, \gls{EEG} sleep staging (Sleep-EDF \citep{kemp2000sleepedf, goldberger2000physionet}, fixed raw sample, greedy, max 80 new tokens).
  The \emph{text} and \emph{soft} columns report only the time-series tokens.
  The full prompt adds a fixed $\sim$211-token pre/post scaffold shared by both arms.
  ``n/a'' marks a series whose \emph{full} prompt exceeds the 4,300-token on-device budget and is skipped to avoid an uncatchable \gls{OS} memory termination.
  \Gls{TTFT} is mean $\pm$ SD over five sweeps; peak memory varied by $\le 3$\,MiB across runs and is reported as the mean; soft-prompt tokens are injected embeddings, not vocabulary tokens.}
  \label{tab:economics}
  \begin{tabular}{r rr r rr rr}
    \toprule
    & \multicolumn{2}{c}{time-series tokens} & & \multicolumn{2}{c}{\gls{TTFT} (s)} & \multicolumn{2}{c}{peak mem (MiB)} \\
    \cmidrule(lr){2-3}\cmidrule(lr){5-6}\cmidrule(lr){7-8}
    $L$ & text & soft & ratio & text & soft & text & soft \\
    \midrule
    100  & 300  & 25  & 12.0$\times$ & $2.15 \pm 0.69$  & $1.29 \pm 0.40$ & 3878 &3621 \\
    250  & 750  & 63  & 11.9$\times$ & $3.88 \pm 1.05$  & $1.46 \pm 0.49$ & 3890 & 3668 \\
    500  & 1500 & 125 & 12.0$\times$ & $6.80 \pm 1.00$  & $1.68 \pm 0.53$ & 4023 & 3722 \\
    1000 & 3000 & 250 & 12.0$\times$ & $14.84 \pm 0.76$ & $2.44 \pm 0.11$ & 4232 & 3836 \\
    1500 & 4500 & 375 & 12.0$\times$ & n/a              & $3.02 \pm 0.10$ & n/a  & 3885 \\
    \bottomrule
  \end{tabular}
\end{table}

\begin{table}[htbp]
  \centering
  \footnotesize
  \caption{On-device context economics, 12-lead \gls{ECG} (\gls{ECGQA} \gls{COT} \citep{oh2023ecgqa}, fixed bundled sample, greedy, max 80 new tokens).
  Columns as in Table~\ref{tab:economics} (time-series tokens; full prompt adds a fixed $\sim$533-token scaffold).}
  \label{tab:economics_ecg}
  \begin{tabular}{r rr r rr rr}
    \toprule
    & \multicolumn{2}{c}{time-series tokens} & & \multicolumn{2}{c}{\gls{TTFT} (s)} & \multicolumn{2}{c}{peak mem (MiB)} \\
    \cmidrule(lr){2-3}\cmidrule(lr){5-6}\cmidrule(lr){7-8}
    $L$ & text & soft & ratio & text & soft & text & soft \\
    \midrule
    100  & 3600  & 300  & 12.0$\times$ & $13.64 \pm 2.58$ & $2.85 \pm 0.77$ & 4372 & 3885 \\
    250  & 9000  & 756  & 11.9$\times$ & n/a              & $4.27 \pm 0.98$ & n/a  & 3959 \\
    500  & 18000 & 1500 & 12.0$\times$ & n/a              & $6.08 \pm 1.14$ & n/a  & 4087 \\
    1000 & 36000 & 3000 & 12.0$\times$ & n/a              & $11.62 \pm 2.16$ & n/a  & 4256 \\
    \bottomrule
  \end{tabular}
\end{table}

\section{Limitations and Future Evaluation}
\label{sec:limitations}

The present evaluation establishes implementation parity and on-device context economics.
The \gls{OPENTSLM} study included clinician review of \gls{ECG} rationales from its model family \citep{opentslm}, and our numerical parity protocol (\S\ref{sec:method}) confirms identical greedy-decoded token sequences between the Swift port and the Python \gls{MLX} reference.
Decoded answers also agree with the PyTorch reference, although rationale wording can differ under \gls{BF16} arithmetic.
This supports implementation equivalence for the evaluated prompts, but it does not constitute new clinical validation.
The benchmarks use fixed Sleep-EDF sleep-staging samples \citep{kemp2000sleepedf, goldberger2000physionet} and \gls{ECGQA} samples \citep{oh2023ecgqa} rather than prospective HealthKit data from patients.
Consumer wearable time-series data may include sensor noise, lower or variable resolution, missing data, device-specific preprocessing, and population shift not captured by these samples.
Prospective studies with real-world wearable time-series data, clinical review, \gls{IRB} approval, and predefined safety boundaries are needed before patient-facing or clinical use.

On-device deployment remains bounded by measured memory and effective prompt budget.
We use Llama-3.2-1B, as it fits the target device while supporting the adapted \gls{OPENTSLMSP} pipeline.
Larger backbones were not evaluated and may exceed the memory budget of current smartphones in this architecture.
Quantization and future hardware may increase feasible model and input size, but the present claim is limited to the 1B backbone, the evaluated device class, and the soft-prompt path tested here.
Tablets such as the iPad Pro may relax the thermal and memory constraints observed on the phone.
The evaluation also covers only the soft-prompt fusion pathway.
Cross-attention fusion, as in Flamingo~\citep{alayrac2022} and OpenTSLM-Flamingo~\citep{opentslm}, is an alternative embedding-level interface, but it modifies the decoder's internal attention layers rather than only the input-embedding sequence adapted here; porting and benchmarking it on device is left to future work.

Local inference reduces exposure from routine cloud inference, but it does not remove governance obligations.
In the current demo, inference runs on the device and HealthKit data are not transmitted during inference.
Whether \gls{HIPAA}, \gls{GDPR}, or other privacy requirements apply depends on the deployment context, data controller or covered-entity role, consent model, logging, export behavior, and downstream data flows.
The local design also changes monitoring trade-offs.
If deployed without central telemetry, performance surveillance and drift detection would require privacy-preserving reporting, local audit mechanisms, or supervised clinical review rather than server-side logs.
Pre-deployment parity testing can confirm that the port preserves reference behavior before evaluation, but it is not a substitute for real-world monitoring.

\section{Conclusion}
\label{sec:conclusion}

PocketTSLM demonstrates that a soft-prompt physiological \gls{TSLM} can run fully on a consumer smartphone in the evaluated setting.
By porting \gls{OPENTSLMSP} to \gls{MLXSWIFT} and integrating it into a HealthKit-enabled demo app, the encoder, projector, and \gls{LLM} execute locally rather than through cloud inference.
The evaluation compares the Swift \gls{MLX} port with the reference implementation under greedy decoding and benchmarks soft prompts against text serialization on fixed Sleep-EDF \citep{kemp2000sleepedf, goldberger2000physionet} and \gls{ECGQA} samples \citep{oh2023ecgqa}.
Soft prompts use approximately $12\times$ fewer time-series tokens than the text-serialized baseline, reducing prompt-prefill cost and allowing longer physiological time-series windows to fit within the measured on-device budget.
These results support PocketTSLM as an implementation and deployment contribution, not as evidence of clinical performance in new settings.
The main remaining step is a prospective evaluation on real-world wearable time-series data with predefined review, safety, monitoring, and governance procedures.

\section*{Ethics Statement}
\label{sec:declarations}

This study uses public, de-identified research datasets and does not introduce new protected health information collection.
No \gls{IRB} approval was obtained for this deployment and benchmarking study, as the reported evaluation did not involve prospective recruitment, participant interaction, intervention, or collection of identifiable private information.

\section*{Acknowledgments}
This manuscript was prepared using \emph{A Reproducible Workflow for Evidence-Centered Scientific Papers}~\citep{schmiedmayer2026papertemplate}.

\clearpage
\bibliographystyle{\paperbibliographystyle}
\bibliography{references}

\end{document}